# Neural translation and automated recognition of ICD10 medical entities from natural language

## Authors

Louis Falissard (corresponding author)

CépiDc Inserm, Paris Saclay University, le Kremlin Bicêtre, France
Postal address: 31 rue du Général Leclerc, 94270, Le Kremlin Bicêtre, France
Email address: louis.falissard@inserm.fr
Phone number: +33679649178

Claire Morgand: CépiDc, Inserm, Le Kremlin Bicêtre, France

Claire Imbaud: CépiDc, Inserm, Le Kremlin Bicêtre, France

Walid Ghosn: CépiDc, Inserm, Le Kremlin Bicêtre, France

Karim Bounebache: CépiDc, Inserm, Le Kremlin Bicêtre, France

Grégoire Rey: CépiDc, Inserm, Le Kremlin Bicêtre, France

Word count : 4200

## Abstract

### Background

The recognition of medical entities from natural language is an ubiquitous problem in the medical field, with applications ranging from medical act coding to the analysis of electronic health data for public health. It is however a complex task usually requiring human expert intervention, thus making it expansive and time consuming.

The recent advances in artificial intelligence, specifically the raise of deep learning methods, has enabled computers to make efficient decisions on a number of complex problems, with the notable example of neural sequence models and their powerful applications in natural language processing. They however require a considerable amount of data to learn from, which is typically their main limiting factor. However, the CépiDc stores an exhaustive database of death certificates at the French national scale, amounting to several millions of natural language examples provided with their associated human coded medical entities available to the machine learning practitioner. This article investigates the applications of deep neural sequence models to the medical entity recognition from natural language problem.

### Methods

The investigated dataset is based on every French death certificate from 2011 to 2016, containing information such as the subject's age, gender, and the chain of events leading to his or her death both in French and encoded as ICD10 medical entities, for a total of around 3 million observations. The task of automatically recognizing ICD10 medical entities from the French natural language based chain of event is then formulated as a type of predictive modelling problem known as a sequence-to-sequence modelling problem. A deep neural network based model known as the Transformer is then slightly adapted and fit to the dataset. Its performance is then assessed on an exterior dataset and compared to the current state of the art. Confidence intervals for derived measurements are derived via bootstrap.

### Results

The proposed approach resulted in a test F-measure of .952 [.946, .957], which constitutes a significant improvement on the current state of the art and its previously reported 82.5 F-measure assessed on a comparable dataset. Such an improvement opens a whole field of new applications, from nosologist level automated coding to temporal harmonization of death statistics.

### Conclusion

This article shows that deep artificial neural network can directly learn from voluminous datasets complex relationships between natural language and medical entities, without any explicit prior knowledge. Although not entirely free from mistakes, the derived model constitutes a powerful tool for automated coding of medical entities from medical language with promising potential applications.

*Keywords*: machine learning, deep learning, machine translation, mortality statistics, automated medical entity recognition, ICD10 coding

# 1 Introduction

The democratization of electronic health record databases has opened countless opportunities to gain precious insights in fields ranging from precision medicine to public health and epidemiology. However, they still present many challenges, both technical and methodological, that make their exploitation cumbersome. As an example, natural language is extensively present in some health related databases, while being notoriously difficult to handle with traditional statistical methods, and preventing most international comparisons due to language barrier. In order to counter these undesirable properties, several approaches have been devised. For instance, by encapsulating most medical entities in a standardized hierarchical tree structure, the ICD10 classification [1] offers a powerful and expressive way of organizing analytics compatible health databases. On the other hand, ICD10 entities are significantly less intuitive for human users than natural language, and require years of training and practice to handle fluently. As a consequence, the data production of classification based medical data is usually handmade, expansive and time consuming. Several attempts have been made to design artificial intelligence based systems able to automatically derive medical entities from natural languages, some with quite promising performance[2]–[4]. However, all of them fall short in automating the complex production schemes inherent to medical databases, specifically in regard to their high data quality standards.

However, recent innovation in deep artificial neural networks have achieved significant progress in natural language processing [5], [6]. In particular, their applications in the field of machine translation[7]–[9], fuelled by increases in both data and computing power, repeatedly bring automated systems closer and closer to human level performances. Several attempts have been made to apply these powerful techniques in an electronic health database setting, most of them with mitigated success. As an example, the current state of the art in ICD10 entity recognition from natural language in death certificates still remains a combination of expert system and SVM based classical machine learning[2]. Several explanations exist for this discrepancy between traditional machine translation and medical entity recognition. First, deep artificial neural network based methods are known to require huge amount of data for optimal performances. However, most experiment were either performed with slightly out-of-date neural architectures, or with dataset sizes at least an order of magnitude under what would be typically required[10]. On the other hand, the "Centre for Epidemiology on Medical Causes of Death" (CépiDc) has been storing French death certificates at the national scale since 2011 in both natural language and ICD10 converted format. The entire database amounts to just under 3 million death examples, thus providing with sensibly better settings to investigate the potential applications of deep neural networks in medical entities recognition.

The following article formulates the process of ICD10 entity recognition from natural language as a sequence to sequence statistical modelling problem (better known as seq2seq models in the academic literature) and proposes to solve it with a variation one of the state of the art machine translation neural architecture, the Transformer. The following section focuses on describing the aforementioned statistical modelling problem and overall methodology. Section 3 reports the result of experiments done on the French CépiDc dataset as well as a comparison with the current state of the art. Section 4 presents a discussion on the model's potential limitation through an error analysis and describes potential leads for improvement.

## 2 Material and methods

### 2.1 Material

The dataset used during this study consists of every available death certificate found in the CépiDc database for the years 2011 to 2016, representing just under 3 million training examples. These documents record various information about their subjects, including the chain of events leading to the subject's death, written by a medical practitioner.

### 2.1.1 Causal chain of death

The causal chain of death constitutes the main source of information available on a death certificate in order to devise mortality statistics. It typically sums up the sequence of events that led to the subject's death, starting from immediate causes (such as cardiac arrest) and progressively expanding into the individual's past to the underlying causes of death. WHO provides countries with a standardized causal chain of events format, which France follows, alongside most developed countries. This WHO standard asks of the medical practitioner in charge of reporting the events leading to the subject's passing to fill out a two-part form in natural language. The first part is comprised of 4 lines, in which the practitioner is asked to report the chain of events, from immediate to underlying cause, in inverse causal order (immediate causes are reported on the first lines, and underlying causes on the last lines). Although 4 lines are available for reporting, they need not all be filled. In fact, the last available lines are rarely used by the practitioner. The second part is comprised of two lines in which the practitioner is asked to report "any other significant conditions contributing to death but not related to the disease or condition causing it" [11] that the subject may have been suffering from.

In order to counter the language dependent variability of death certificates across countries, a pre-processing step is typically applied to the causal chain of events leading to the individual's death, where each natural language based line on the certificate is converted into a sequence of codes defined by the 10th revision of the International Statistical Classification of Diseases and Related Health Problems (ICD-10)[1]. ICD-10 is a medical classification created by WHO defining 14199 medical entities (e.g. diseases, signs and symptoms…) distributed over 22 chapters and encoded with 3 or 4 alpha decimal symbols (one letter and 2 or 3 digits), 5615 of which are present in the investigated dataset. Table 1 shows an example of a causal chain of events taken from an American death certificate, in both natural language and ICD10 formats.

| Line | Natural language | ICD10 encoding |
|---|---|---|
| 1 | STROKE IN SEPTEMBER LEFT HEMIPARESIS | I64 G819 |
| 2 | FALL SCALP LACERATION FRACTURE HUMERUS | S010 W19 S423 |
| 3 | CORONARY ARTERY DISEASE | I251 |
| 4 | ACUTE INTRACRANIAL HEMORRHAGE | I629 |
| 6 | DEMENTIA DEPRESSION HYPERTENSION | F03 F329 I10 |

*Table 1: Example of cause chain of death, in natural language and as ICD10 codes. Some natural language lines correspond to several ICD10 codes, whose orders matter in the overall coding process*

As aforementioned, the process of converting the natural language based causal chain of events leading to death in an ICD10 format is the main focus of this article. Consequently, the latter will be selected as target variable and the former as the main explanatory variable for the neural network based predictive model defined further.

For reasons related to the underlying cause of death production process, the natural language based chain of events and its ICD10 encoded counterpart suffer from alignment errors at the line level, as shown in table 2. Although qualitatively deemed quite rare, this misalignment phenomenon brings sufficient noise in the dataset to prevent model convergence while fitting models with line level sentence pairs.

| Line | Natural language | ICD10 encoding |
|---|---|---|
| 1 | STROKE IN SEPTEMBER LEFT HEMIPARESIS | I64 G819 |
| 2 | FALL SCALP LACERATION FRACTURE HUMERUS | S010 W19 S423 |
| 3 | CORONARY ARTERY DISEASE | I629 I251 |
| 4 | ACUTE INTRACRANIAL HEMORRHAGE | |
| 6 | DEMENTIA DEPRESSION HYPERTENSION | F03 F329 I10 |

*Table 1: Same certificate as displayed in table 1 showcasing the misalignment phenomenon. The ICD10 code related to line 4 (both in red) has been moved to line 3 by a human coder. Concatenating lines in a backward fashion restores alignment while preserving ordering*

In order to bypass this critical flaw in the investigated dataset, a decision was chosen to consider as input and target variables the certificates lines concatenated in a backward fashion (from line 6 to line 1), as can be seen in figure 1. This slight change in data format does not significantly alter the problematic at hand, as the investigated model is still trained to recognize ICD10 encoded medical entities from natural language. If anything, the modified modelling problem can be expected to be more difficult, as both the variance and dimensionality of both input and target variables have increased. Several methods are available to retrieve line level aligned predictions from a model trained in such a configuration, for instance using a combination of transfer learning and pruned tree search.

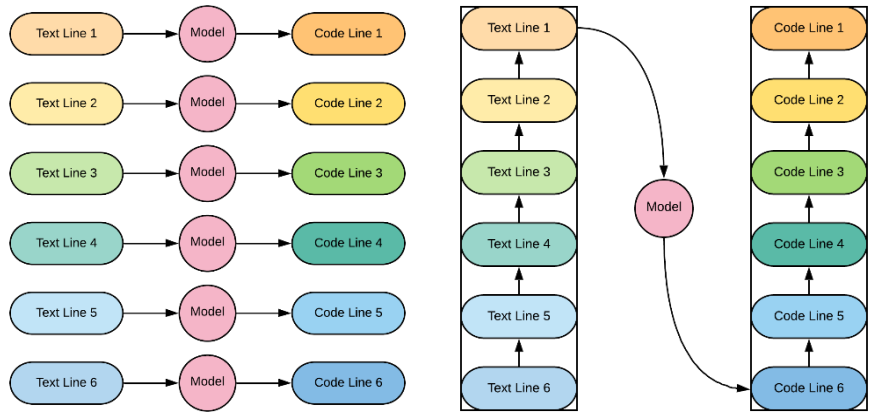

*Fig. 1 Left: the original modelling problem. Each certificate line is taken as an input variable to predict its corresponding ICD10 code line. Right: The modified investigated problem. All certificate lines are concatenated and taken as an input variable to predict the corresponding concatenated ICD10 code line*

### 2.1.2 Miscellaneous variables

From gender to place of birth, a death certificate contains various additional information on its subject besides the chain of events leading to death. As some of these items are typically used by both expert systems and human coders to detect ICD10 entities in the chain of events, they present an interest as explanatory variable for the investigated predictive model. After consultation with expert coders, the following items available on French death certificate were selected as additional exogenous variables:

- gender (2 states categorical variables),
- year of death (6 states categorical variables),
- age, factorized into 5 years' intervals with the exception of subject less than one-year-old, which were divided into two classes following whether they were more than 28-day-old,
- origin of the death certificate (2 states categorical variables, either from the electronic or paper based death certification pipeline).

Strictly speaking, the subject's year of passing should only have a limited effect on the relationship between natural language and its contained medical entities. However, the WHO defined coding rules, as well as their interpretations by human coders slightly evolve over the years. As a consequence, the model should benefit, in term of predictive performance, from being able to differentiate between different years.

Similarly, the impact of the certificate's origin on the model's predictive power is not entirely obvious at first sight. However, the paper based certificates data entry process is handled by human through speech recognition technology. In addition, the entry clerks are asked to apply a small set of normalization rules to the natural language. Electronic death certificates, however, are received directly from the medical practitioner as is. As a consequence, distribution shifts are to be expected from paper to electronic based chain of events, and including this information as an explanatory variable might be beneficial to the model's predictive power.

## 2.2   Method

With both the explanatory and target variables well defined, the investigated modelling problem can be defined as follows:

$$P(ICD|NL, A, G, Y, E, \theta) = f_\theta(NL, A, Y, G, E)$$

With:

- $ICD \in [\![0, 1]\!]^{5616^{20}}$ One line of a certificate encoded as ICD10 entities
- $NL \in [\![0, 1]\!]^{V^L}$ the line in natural language, tokenized with a vocabulary $V$ and of maximum sequence length $L$
- $A \in [\![0, 1]\!]^{25}$ the categorized age
- $Y \in [\![0, 1]\!]^{6}$ the year of death
- $[\![0, 1]\!] \in \mathbb{R}^2$ the gender
- $[\![0, 1]\!] \in \mathbb{R}^2$ the death certificate's origin
- $f_\theta$ a mapping from the problem's input space to its output space, parameterized in $\theta$ a real-valued vector (typically a neural network)

Theoretically, the derived modelling problem is typical of traditional statistical modelling problems, and could be solved using multinomial logistic regression. In practice, however, this approach presents a significant drawback. In this setting, the investigated target variable constitutes a categorical variable with $5616^{20}$ (20 ICD10 codes sequences, each of which can take 5616 values) distinct states, thus rendering the analysis untractable both in term of computational expanses and sample size requirements. This type of approach, however, makes no use of the data's inherent sequential nature, which allows to rewrite the investigated modelling problem as follows:

$$P(ICD|NL,A,G,Y,E,\theta) = P\big((ICD_1,-,ICD_n)\big|NL,A,G,Y,E,\theta\big) \; \forall n \in [\![1,20]\!]$$

$$= \prod_{i=1}^{n} P(ICD_i|(ICD_1,-,ICD_{i-1}),NL,A,G,Y,E,\theta) \; \forall n \in [\![1,20]\!]$$

With:

- $ICD_i \in [\![0,1]\!]^{5616} \; \forall i \in [\![1,20]\!]$ the i$^{th}$ code present on the code line

Factors in the right hand side of equation 2 can be interpreted as distinct predictive modelling problem, all with an output variable distributed across all ICD10 codes. Although still highly dimensional, predicting output variables of such dimensionality is typically tractable with modern machine learning techniques[7]. They present however two significant drawbacks for traditional modelling techniques:

- The number of output variables to predict varies across observations in the dataset (not all death certificates have 20 ICD10 codes)
- The output variables' distributions are conditioned on previous ones

This particular formulation is known in the deep artificial neural network community as a sequence to sequence modelling problem[7], and has been an active area of research for the past few years. As one of the state of the art neural architecture devised in the field, the Transformer[9] was chosen as the predictive model investigated in the following experiments. It was recently outperformed by the Evolved Transformer[12], a variation on the former. However, both approaches were investigated and yielded similar results. The Transformer architecture was retained due to its availability of official and maintained implementations, and the final results further displayed were obtained using an ensemble of 7 such models.

Several specificities in the aforedefined modelling problem required small adaptations to the Transformer architecture. However, the authors feel their technicity fall outside the scope of this article. The interested reader will however find a complete description of these modifications in the annex documents.

### 2.3 Training and evaluation methodology

The investigated model was trained using all French death certificates from years 2011 to 2016. 5000 certificates were randomly excluded from each year and distributed into a validation set for hyper-parameter fine-tuning, and a test dataset for unbiased prediction performance estimation (2500 each), resulting in three datasets with following sample sizes:

- Training dataset: 3240109 records
- Validation and test dataset: 30000 records each

The model was adapted from Tensorflow's (a python-based distributed machine learning framework) official Transformer implementation. Training was performed on three NVidia RTX 2070 GPUs simultaneously using a mirrored distribution strategy using a variant of stochastic gradient descent, the Adam optimization algorithm.

Hyper-parameters were first initialized following the Transformer's authors in their base setting. Further fine tuning of a selected number of hyper-parameters was performed using a random search guided on the validation set. The interested reader will find a complete description of the training process and hyper-parameter values defining this model in annex.

After training, the model's predictive performance was assessed on the test dataset (excluded prior to training, as mentioned earlier), and compared to the current state of the art, obtained by the "Laboratoire d'Informatique et de Mécanique pour les Sciences de L'ingénieur" (LIMSI) during the 2017

CLEF eHealth challenge[2]. As the CLEF eHealth challenge only provided electronic certificates to the contestants, and in order to ensure comparability, the model's performances were assessed on paper and electronic certificates separately. For the same reason, the performance metrics used for model evaluation were selected as follows:

$$Precision = \frac{True\ positives}{True\ positives + False\ positives}$$

$$Recall = \frac{True\ positives}{True\ positives + False\ negatives}$$

$$F\text{-}measure = \frac{2 \cdot Precision \cdot Recall}{Precision + Recall}$$

With:

- True positives the number of codes predicted by the model that are present in the test set's true output target,
- False positives the number of codes predicted by the model that are not present in the test set's true output target,
- False negatives the number of codes not predicted by the model that are present in the test set's true output.

The informed reader might find these metrics stray away from common machine translation system benchmarking metrics such as BLEU or negative log perplexity scores[7]–[9], [13], but the former were the only ones used in comparable work. As BLEU and negative log perplexity have close to no absolute interpretability without comparisons to alternative methods, their use was discarded from the experiment. In order to present the reader with a more comprehensive view of the proposed approaches' performances, these accuracy metrics were also derived on a per chapter basis, again on the same test set, and confidence intervals were computed using bootstrap.

## 3   Results

The ensemble of transformer models were trained as aforedescribed for approximately 3 weeks, and the final ensemble's predictive performance as well as the current state of the arts' are reported in Table 1. As previously mentioned, the current state of the arts' performances were assessed on electronic certificates only, and should as a consequence be compared to the proposed approach performance on a similar situation. Because paper based certificates are still sensibly more common than their electronic counterparts in France (approximately 90% of certificates in the dataset are paper based), overall and paper specific performances are also displayed.

| Approach | F-measure | Precision | Recall |
| --- | --- | --- | --- |
| Current state of the art (LIMSI) | .825 | .872 | .784 |
| Proposed approach (electronic certificates) | **.952 [.946, .957]** | **.955 [.95, .96]** | **.948 [.943, .954]** |
| Proposed approach (paper certificates) | .942 [.941, .944] | .949 [.947, .95] | .936 [.934, .937] |
| Proposed approach (all certificates) | .943 [.941, .944] | .949 [.948, .951] | .937 [.935, .938] |

Table 3 F-measure of the current state of the art and the proposed approach, with their corresponding 95% confidence intervals, derived by bootstrap. Confidence intervals were not provided in the LIMSI's publication and are therefore not displayed.

The proposed approach shows an F measure 73% closer to a perfect score when compared to the current state of the art. In addition to its substantial improvement in F-measure, the proposed approach displays significantly more balanced precision and recall scores than the LIMSI's method (from 5% relative difference to less than 1%).

A surprising result, however, lies in the model's lower performances on paper certificates. Indeed, the standardization they receive due to their voice based data collecting process considerably reduces variance and prevents any misspelling of words in the data potentially present in electronic based certificates. As a consequence, model performance on the former should be expected to be higher. A potential explanation for this phenomenon lies in the potential for missing data in paper based certificates. Indeed, when confronted to poorly written words, data clerks are allowed to replace them with a "!" symbol when the word is estimated unreadable (present in approximately 10% of paper based certificates). Medical coders, however, are usually more efficient in guessing the words from the written certificates (typically with the addition of contextual clues). A purely text based approach however, is then limited to pure guess on those observations with missing data, logically leading to poorer performance. This phenomenon being absent from electronic based certificates, it constitutes a promising candidate in explaining this unexpected difference of performance. In addition, model performances on paper certificates not containing any "!" symbol in the test set led to 96.2% F-measure, thus providing strong evidence to support this hypothesis.

## 4 Discussion

Although the proposed approach significantly outperforms the current state of the art, neural network based methods are known to present several drawbacks that can significantly limit their application in some situations. Typically, the current lack of systematic methods to interpret and understand neural network based model and their decision processes can lead the former to perform catastrophically on ill predicted cases, independently from their high predictive performances. As a consequence, the proposed model behaviour in ill predicted cases require careful analysis. In addition, such an investigation can lead to significant insights potentially relevant when applying the derived model in practical applications.

### 4.1 Per-chapter quantitative analysis

One simple, straightforward approach to understanding the model's weakness, lies in assessing its performance on a finer grain level, for instance by identifying false positives and negatives not only at the global level, but per ICD10 chapters, as can be seen in table 4.

It appears from these graphs that although the most prevalent medical entities are associated with low false positive and negative rates, some rarer chapters are associated with unreasonably high error rates. Depending on their prevalence and accuracies, these chapters can be classified into two distinct categories:

- Chapters associated with unreasonably high error rates but extremely low prevalence such as "diseases for the ear and mastoid process" or "pregnancy, childbirth and the puerperium". However, these entity groups remain rare enough within the dataset to allow for alternative treatments, like manual evaluation, for instance.
- Chapters associated with high error rates (although lower than the former) but with significant prevalence such as "External causes of morbidity and mortality" or "Injury, poisoning and certain other consequences of external causes".

The task of identifying these potential mistakes, however, is not entirely trivial depending on whether mistakes are of false positive or false negative types. Indeed, potential false positives errors are directly identifiable within the predicted ICD10 code sequences. As a consequence, coding quality control for this mistake type should be fairly straightforward to implement (one could for instance manually review all code sequences containing codes related to "Pregnancy, childbirth and the puerperium" systematically. Potential false negative errors, however, are inherently significantly harder to identify, and require further investigation, for instance through association rules analysis.

A number of promising leads are already available and should reasonably improve upon the proposed approach:

- Training methods adapted to imbalanced datasets such as up sampling or loss weighting
- Data augmentation for rare medical entities
- Addition of information to the model (prenatal related death, for instance, are explicitly defined as such on certificates)
- Hybrid approach with traditional NLP approaches, typically less expensive in term of sample size requirements

| ICD10 chapter | False positives (%) | False negatives (%) | Prevalence (%) |
|---|---|---|---|
| Diseases of the circulatory system | 3.75 | 4.98 | 22.4 |
| Symptoms, signs and abnormal clinical and laboratory findings, not elsewhere classified | 3.87 | 4.12 | 21.8 |
| Neoplasms | 4.07 | 5.07 | 15.9 |
| Diseases of the respiratory system | 3.02 | 4.00 | 8.76 |
| Endocrine, nutritional and metabolic diseases | 2.17 | 3.44 | 4.83 |
| Diseases of the nervous system | 2.70 | 4.12 | 3.89 |
| Mental and behavioural disorders | 2.88 | 4.14 | 3.58 |
| Diseases of the digestive system | 5.72 | 8.10 | 3.53 |
| Factors influencing health status and contact with health services | 19.2 | 19.6 | 3.08 |
| Diseases of the genitourinary system | 5.45 | 7.59 | 2.71 |
| External causes of morbidity and mortality | 16.6 | 23.5 | 2.57 |
| Certain infectious and parasitic diseases | 7.98 | 9.23 | 2.55 |
| Injury, poisoning and certain other consequences of external causes | 14.0 | 19.8 | 2.07 |
| Diseases of the blood and blood-forming organs and certain disorders involving the immune mechanism | 6.72 | 12.2 | 0.77 |
| Diseases of the musculoskeletal system and connective tissue | 12.2 | 17.3 | 0.62 |
| Diseases of the skin and subcutaneous tissue | 8.72 | 8.16 | 0.51 |
| Certain conditions originating in the perinatal period | 14.5 | 20.5 | 0.16 |
| Congenital malformations, deformations and chromosomal abnormalities | 22.4 | 25.6 | 0.15 |
| Diseases of the eye and adnexa | 4.93 | 13.6 | 0.076 |
| Codes for special purposes | 24.0 | 34.0 | 0.047 |
| Diseases of the ear and mastoid process | 5.6 | 33.3 | 0.017 |
| Pregnancy, childbirth and the puerperium | 50 | 33.3 | 0.0056 |

Table 4 Prevalence, false positives and negatives rates for each ICD10 chapter, sorted in descending order by prevalence

## 4.2 Score calibration fitness assessment

The model being fit in a similar fashion to multinomial logistic regression, it not only yields a prediction, but an associated score similar to a confidence probability. If properly calibrated, this score can offer powerful insights regarding the prediction's quality at the individual level. Typically, a "good" score would be expected to show higher values in cases where the predicted ICD10 sequence is of good quality (typically in term of F-measure) and a low one when mispredicting. Efficient assessment of such scores in traditional machine learning problems is typically done through visualization of ROC curves. However, the sequential, multinomial nature of the investigated problem renders this approach ill defined. The plot found in figure 2, while conceptually similar to a ROC curve, was derived following a slightly different approach in order to efficiently appreciate the model score's quality:

- A grid of score threshold values was defined (uniform grid with 0.01 intervals)
- For every given threshold value were computed the percentage of predictions with inferior or equal scores (considered as rejected predictions due to poor score), as well as the F-measure performance on the accepted predictions
- Percentage of accepted certificates and F-measurement were scatter plotted against each other, with threshold value displayed as points' colour

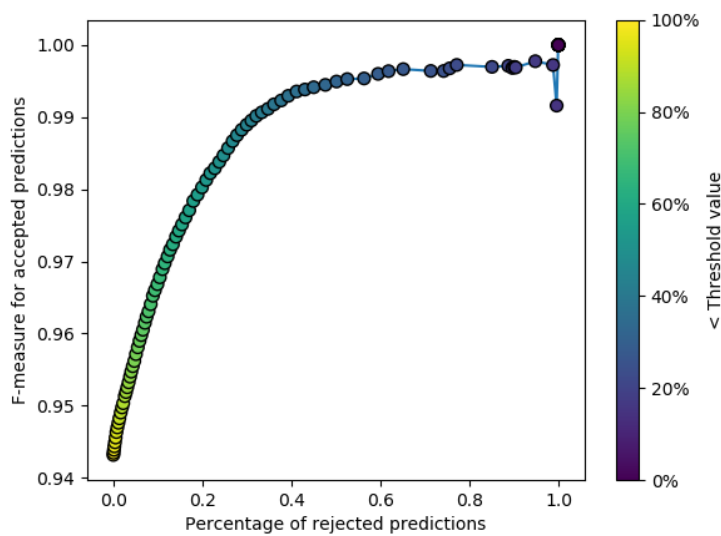

Fig. 2 Percentage of rejected predictions versus F-measure on accepted ones. The score threshold value defining the accepted predictions are displayed as point colours

By showing a clean, increasing relationship between number of rejected predictions and F-measure on the remaining, figure 2 strongly indicates good score calibration. As an example, setting a threshold score of approximately 50% allows for a fully automated coding of 80% of the certificates presents in the test set with a corresponding F-measure exceeding 98%.

### 4.3 Human based qualitative error analysis

The error analysis carried on so far allowed for the assessment of the model's strength and weakness on the global level. They however fail to yield any interesting insight regarding potential model biases, for instance towards specific coding rules. Indeed, the coding of medical entity from natural language, especially with regard to mortality statistics, is subject to a number of coding rules depending on context or pathology, with a level of specificity oftentimes reaching casuistry.

In addition, all results presented so far with a model error defined as a disagreement between the model's output and the information contained in the database. However, building a medical database is a complex, mostly human based process. As such, an inevitable amount of noise is to be expected in the ICD10 coding process, potentially leading to negative bias toward our performance and error evaluations, the aforementioned presence of missing data in the natural language being a perfect example of such phenomenon.

One straightforward, although fairly time-consuming approach to address these two considerations can be derived from human observation of disagreement cases by a ICD10 coding specialist. Two experiments were conducted following this idea.

First, 99 ill predicted certificates were selected at random from the test set and shown to the medical practitioner referent and final decision-maker on ICD10 mortality coding in France, who was asked for each certificates to:

- Extract all the ICD10 medical entities present on each death certificates by herself, from the information the proposed model had access to,
- Give a qualitative comment on the investigated model and database's outputs compared to hers.

The derived ICD10 sequences were then compared to both the actual values contained in the dataset and those predicted by the derived model with the aforedefined metrics used a similarity distance. For better comparability, these statistics are reported both on:

- All certificates in table 5
- Certificates without missing data in the natural language based causal chain of death (by excluding certificates containing a "!" symbol) in table 6

| Database or prediction | F-measure | Precision | Recall |
|---|---|---|---|
| Database against medical expert | .891 [.859, .920] | .868 [.827, .905] | .916 [.888, .940] |
| Prediction against medical expert | .894 [.867, .918] | .894 [.860, .923] | .894 [.868, .920] |

Table 5. F-measure, precision and recall (with their 95% confidence intervals) of both the database and the model's predictions against the medical expert for all sampled certificates

| Database or prediction | F-measure | Precision | Recall |
|---|---|---|---|
| Database against medical expert | .909 [.873, .939] | .901 [.855, .938] | .917 [.888, .940] |
| Prediction against medical expert | .877 [.845, .910] | .877 [.837, .911] | .877 [.847, .906] |

Table 5. F-measure, precision and recall (with their 95% confidence intervals) of both the database and the model's predictions against the medical expert for sampled certificates without missing data

Tables 5 and 6, show no significant difference in prediction performance between the proposed approach and the current data production process (based on a combination of expert system and human coders). When including certificates containing missing text, the proposed model slightly outperforms the current system (although not significantly), further confirming the hypothesis that the performance metrics reported in the result section are negatively biased. When excluding these problematic observations, the expert system/human coder combination slightly outperforms the proposed approach, although still not significantly. The observed drop in the investigated model's performance when excluding missing text certificates might seem surprising at first, as the information available to both the medical practitioner and the model remains the same regarding of this exclusion. However, this performance gap might be explained through selection bias, as faulty predictions were selected by their difference with the database's ICD10 content. The observed F-measure of .974 observed between model prediction and medical expert opinion on certificates with missing text provides strong evidence to confirm this hypothesis.

From the qualitative comments made by the medical experts, three major types of model errors could be defined:

- In 16% of cases, disagreement between the current data production process and the proposed approach was due to missing information in the input text. On these specific cases, the F-measure between model output and medical expert decision was measured at .974 (an example of such error case can be seen in the annex, in table MA4)
- In 14% of cases, the correct ICD10 sequence is dependent on highly contextual cues or external knowledge of world behaviour (e.g. Someone found dead at the bottom of stairs is quite likely to have suffered a fall. An example of such error case can be seen in the annex, in table MA3)
- In 12% of cases, the correct ICD10 sequence is dependent on highly nonlinear, almost casuistic rules and are typical examples of scenarios where a hybridized deep learning and expert-based system should be beneficial (an example of such error case can be seen in the annex, in table MA4)
- The remaining cases did not elicit any comment from the medical expert.

Finally, in a second experiment, the medical expert's ability to discriminate between human coding and the proposed approach was assessed, in a Turing test-like approach. To do so, a hundred additional ill-predicted certificates were sampled at random from the test set, and associated with their two corresponding, anonymized and shuffled ICD10 sequences (from both the proposed approach and the database). The medical expert was then asked, after careful reviewing of each certificate, to answer the question "Which of these ICD10 sequences candidates was derived by the human/expert system combination". After exclusion of certificates containing missing text data (where the human coder was easily identifiable due to the apparently out of context additional codes as seen in table MA2), the medical expert was able to correctly identify the human in 62.0% [50.7, 73.2] of cases, which is significantly better than random guessing (although barely).

# 5 Conclusion

In this article, the task of automatic recognition of ICD10 medical entities from natural language in French was presented as a seq2seq modelling problem, well known in the deep artificial neural network academic literature. From this consideration, the performances of a well-known approach in the field, consisting of an ensemble of Tranformer models, was investigated using the CépiDc database and shown to obtain a new state of the art. The derived model's behaviour was thoroughly assessed following different approaches in order to identify potential weaknesses and leads for improvements. Although the proposed approach significantly outperforms any other existing automated ICD10 recognition systems on French free-text, the question of method transferability to other languages require more investigations.

The substantial performances reported in this article open an entire range of promising applications in various medical related fields, from medical act automated coding to advanced natural language based analysis for epidemiology. However, these interesting opportunities are oftentimes prohibited by these methods' massive drawbacks, mostly their requirement for millions of annoted observations to perform well. Mortality datasets, in spite of their specificity, provide researchers with huge, clean and multilingual medical text data perfectly fit for the application of deep neural networks. As a consequence, and keeping in mind neural network's strong transfer learning capability, the authors firmly believe that mortality data constitutes one of the most promising point of entry into modern natural language processing methods applications in the biomedical sciences.

# Neural translation and automated recognition of ICD10 medical entities from natural language – Annex

## 1 Pre-processing

### 1.1 Text standardization

Minimal standardization was applied to the text data. The 6 lines present on the death certificates were concatenated with a ", " separator, and the two following steps were applied as the only text cleaning treatments:

- All letters were put to lower case
- All space based separator were collapsed to " "

### 1.2 Tokenization for rare words

In order to reduce the problem's dimensionality and handle rare words in the dataset, Byte pair encoding[1], the standard methodology used in the recent machine translation academic literature, was used. Its implementation used for the experiments reported in this article can be found in the official Tensorflow Transformer repository[2].

The algorithm was applied on the entire training dataset and the derived tokenization were of dimensionality 500 and 2033 for the ICD10 and French corpora, respectively.

## 2 Model definition

The model itself follows the traditional transformer architecture[3]. The model's official Tensorflow implementation was used for the experiments[3]. However, the traditional Transformer model doesn't allow for the treatment of additional conditional variables. In order to include the latter in the model, a similar methodology than that followed in [4] was chosen:

- Each conditional variable was linearly projected linearly in an embedding space whose dimensionality match the transformer's hidden size hyper-parameter
- These linear projections are added in an element-wise fashion to the embedded token sequence
- The transformer model is used as defined in its original article on the resulting embedded sequence

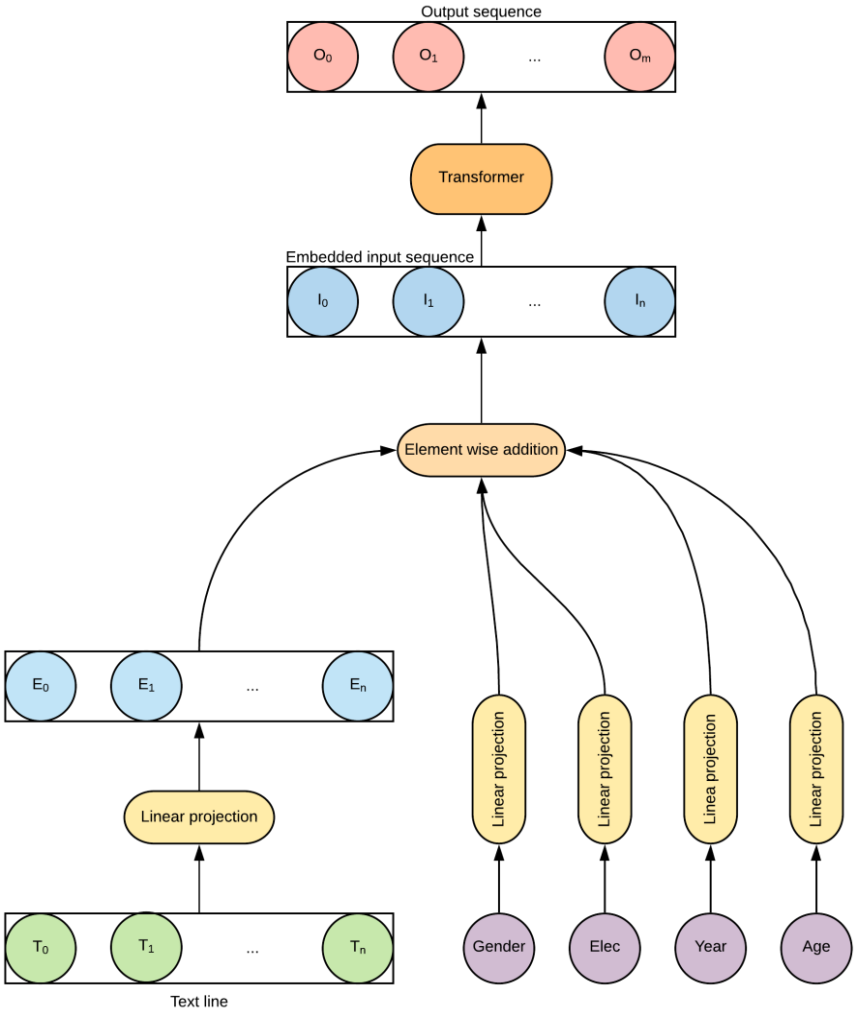

Fig. 1 Transformer adaptation for the handling of conditional variables

# 3   Hyperparameter search

Hyperparameter tuning was done with a random search guided by validation set's F-measure results.

The following variable were randomly sampled from the further specified probability distributions:

- Model's hidden size: sampled from a uniform random distribution between 256 and 512
- Batch size: For computational reasons, the batch size was defined as (100 * 512 / hidden_size)
- Learning rate: Uniformly sampled from discrete values 1. or 2. (note that this value doesn't constitute the actual learning rate, which is modified by the function "get_learning_rate"
- Layer_postprocess_dropout: sampled from a uniform random distribution between 0 and 0.2
- Attention_dropout: sampled from a uniform random distribution between 0 and 0.2
- Relu_dropout: sampled from a uniform random distribution between 0 and 0.2

All other parameters were fixed as recommended by the BASE_MULTI_GPU settings provided in the tensorflow transformer official implementation.

40 models were trained with different hyperparameters sampled from these distributions, the best set of hyperparameter was then used to train a new set of model for ensembling.

# 4   Ensembling method

Due to computational reasons, the traditional method for ensembling neural translation model (logits averaging during the beam search process) could not be used. The following alternative was used instead:

- Get the prediction from each model
- Compute F-measurements between all prediction candidates
- Select the prediction that shows highest F-measurements with other candidates on average

The ensemble of models was selected by a greedy search on all the models trained for the experiment (40 models trained during the hyperparameters search and additional models trained with the best hyperparameter set) guided by the F-measurement on the validation set

The derived score was taken as the prediction scores' average.

# 5 Final ensemble hyperparameters

The final ensemble found by greedy exploration consisted of 7 different models, 5 of which were trained with the best set of hyperparameters revealed by the random hyperparameter searchs. The three distinct sets of hyperparameters can be found in table 1.

| Hyperparameter | Set 1 (best set) | Set 2 | Set 3 |
| --- | --- | --- | --- |
| Batch size | 172 | 152 | 164 |
| Hidden size | 296 | 336 | 312 |
| Learning rate | 2. | 2. | 2. |
| Layer postprocess dropout | .073 | .12 | .005 |
| Attention dropout | .105 | .030 | .017 |
| Relu dropout | .173 | .030 | .20 |

Table MA1. Sets of hyperparameters for the different models used in the final ensemble

# 6   Error examples

| Text | hta, insuffisance cardiaque, anévrisme aorte !, !, asystolie ! |
|---|---|
| Predicted ICD10 | I10 I509 I714 I500 |
| Database ICD10 | I10 I509 I714 H570 I500 R068 |
| True ICD10 | I10 I509 I714 I500 |

Table MA2. Example of "missing data" type error. The database shows two additional codes that are not present in the text according to the medical expert. These codes are probably associated with the "!" present in the text, and were derived from a human coder reading the handwritten death certificate.

| Text | acfa, hta, connu vertige, retrouvé terre bas escalier |
|---|---|
| ICD10 prediction | I48 I10 R42 R98 |
| Database ICD10 | I48 I10 R42 W10 |
| True ICD10 | I48 I10 R42 W10 |

Table MA3. Example of contextual error. The proposed approach converts "retrouvé terre bas escalier" (which roughly translates to "found at the bottom of the stairs") to R98 "unattended death". Both human coders are able to deduce that the subject probably fell down the stairs and use the ICD10 code W10 "Fall on and from steps"

| Text | cardiopathie ischemique avec triple pontage, anevrisme aortique, cancer de la vessie, hta, arret cardio - respiratoir |
|---|---|
| Predicted ICD10 | I259 Z951 I719 C679 I10 R092 |
| Database ICD10 | I259 I251 I719 C679 I10 R092 |
| True ICD10 | I259 I251 I719 C679 I10 R092 |

Table MA4. Example of error caused by a coding rule. I251 and Z951 are both suitable for "triple pontage" (Coronary artery bypass surgery). However, the M4 mortality coding rule (Special instructions on surgery and other medical procedures) dictates the code choice